\pdfoutput=1

\documentclass[11pt]{article}

\usepackage{acl}

\usepackage{times}
\usepackage{latexsym}

\usepackage[T1]{fontenc}

\usepackage[utf8]{inputenc}

\usepackage{microtype}

\usepackage{url}

\usepackage{tabularx}
\usepackage{graphicx}
\usepackage{booktabs}
\usepackage{multirow}
\usepackage{multicol}
\usepackage{paralist}
\usepackage{inconsolata}

\usepackage{xcolor}

\usepackage{xspace}

\newcommand{\unverif}{\textsc{unverifiable}\xspace}
\newcommand{\supp}{\textsc{supported}\xspace}
\newcommand{\refu}{\textsc{refuted}\xspace}

\newcommand{\bear}{\textsc{bear} \citep{wuhrl-klinger-2022-recovering}\xspace}
\newcommand{\bearfact}{\textsc{bear-fact}\xspace}
\newcommand{\F}{F$_1$\xspace}

\usepackage{tikz}
\usetikzlibrary{shapes.geometric}
\usepackage{caption}
\usepackage{subcaption}

\usepackage{hyperref}
\usepackage{float}

\renewcommand{\paragraph}[1]{\par\noindent\textbf{#1}}

\title{What Makes Medical Claims (Un)Verifiable?\\Analyzing Entity and Relation Properties for Fact Verification}

\author{Amelie W\"uhrl$^1$, Yarik Menchaca Resendiz$^1$, Lara Grimminger$^1$, \and Roman Klinger$^{1,2}$ \\
  $^1$Institut f{\"u}r Maschinelle Sprachverarbeitung, University of
  Stuttgart, Germany \\
  $^2$Foundations of Natural Language Processing, University of Bamberg, Germany \\
  \texttt{\{amelie.wuehrl,
    yarik.menchaca-resendiz,}\\\texttt{lara.grimminger\}@ims.uni-stuttgart.de}\\
  \texttt{roman.klinger@uni-bamberg.de}
}

\begin{document}
\maketitle
\begin{abstract}
Biomedical claim verification fails if no evidence can be discovered. In these cases, the fact-checking verdict remains unknown and the claim is unverifiable. To improve upon this, we have to understand if there are any claim properties that impact its verifiability. In this work we assume that entities and relations define the core variables in a biomedical claim's anatomy and analyze if their properties help us to differentiate verifiable from unverifiable claims. In a study with trained annotation experts we prompt them to find evidence for biomedical claims, and observe how they refine search queries for their evidence search. This leads to the first corpus for scientific fact verification annotated with subject--relation--object triplets, evidence documents, and fact-checking verdicts (the \textsc{bear-fact} corpus). We find (1) that discovering evidence for negated claims (e.g., X--does-not-cause--Y) is particularly challenging. Further, we see that annotators process queries mostly by adding constraints to the search and by normalizing entities to canonical names. (2) We compare our in-house annotations with a small crowdsourcing setting where we employ medical experts and laypeople. We find that domain expertise does not have a substantial effect on the reliability of annotations. Finally, (3), we demonstrate that it is possible to reliably estimate the success of evidence retrieval purely from the claim text~(.82\F), whereas identifying unverifiable claims proves more challenging (.27\F). 
  The dataset is available at \url{http://www.ims.uni-stuttgart.de/data/bioclaim}. 

\end{abstract}

\section{Introduction}
Verifying scientific claims in user-generated content is sometimes unsuccessful, 
because no supporting or refuting evidence can be found. In these cases, the claim remains unverifiable. Previous work shows that both evidence retrieval and claim verification, the two core steps in automatic fact verification, are more robust for concisely formulated claims that have been extracted from noisy context compared to verifying user-generated claims directly~\citep{sundriyal-etal-2022-empowering,wuehrl-etal-2023-entity}.

Based on these findings, we hypothesize that breaking down claims into smaller units increases our understanding which properties impact verifiability. This knowledge is key to improving fact-checking (FC) systems.  
To this end, we assume that biomedical entities, e.g., mentions of treatments or medical conditions, and the relations between them (\textit{causes}, \textit{is--a--side-effect} etc.) make up the core variables that define the claim's anatomy. We analyze if these variables are connected to the claim's verifiability, i.e., that there are reoccurring patterns with respect to which types of claims tend to be \supp, \refu, or \unverif.

While disciplines like argument mining have an evolved understanding of claim properties \citep{boland2022}, biomedical claims are poorly understood. The data resources to perform analyses do not exist yet: there is no corpus which is annotated both with (i) biomedical entities in relation that constitute claims and (ii) evidence and the veracity
label it leads to. To create such a resource, we perform (1a) an annotation study of
medical tweets in which we observe carefully trained in-house
annotator's behavior to find evidence for entity-centered
claims, (1b) a statistical analysis of the connection between
entities/relations and the successful evidence retrieval, (2) a
comparison of the in-house annotators' performance to 
crowdsourcing, in which we task laypeople and medical experts to verify the same claims. Finally, we (3) compare the performance of a fine-tuned transformer model to estimate the
checkability of a claim.

We contribute \bearfact, a novel Twitter\footnote{Twitter is now called X.} dataset for biomedical fact verification. It consists of 1,448 fact-checked claims, evidence documents and structured entity/relation
information. We answer the following research questions:

\begin{compactitem}
\item[\textbf{RQ1a}]Which properties, i.e., entity-relation patterns, make a
  claim (un)verifiable?
\item[\textbf{RQ1b}]How can we use medical entities in the claims as meaningful
  search queries for evidence discovery?
\item[\textbf{RQ2}]What is the impact of the annotation setting, i.e., domain
  knowledge and crowdsourcing on annotation quality?
\item[\textbf{RQ3}]Can we predict the verifiability, i.e., the likelihood that evidence for a claim exists, purely from
  the claim?
\end{compactitem}

For RQ1a (§\ref{claim-characteristics-analysis}), we find that entity--relation patterns in claims are
connected to verifiability. Claims conveying a positive relation
(e.g., \textit{cause--of}) are more successfully fact-checked and more
frequently \supp compared to their negative counterpart (\textit{not--cause--of}). For RQ1b (§\ref{evidence-discovery-analysis}), we find that study participants
predominantly change an entity-centric predefined query by
reformulating entity realizations to canonical names. For RQ2 (§\ref{crowd-settings}), we observe that domain expertise
does not have a substantial effect on the reliability of fact-checking
annotations. Finally, the hypothesis in RQ3 holds to some degree (§\ref{modeling}): Fine-tuning a RoBERTa model to differentiate between verifiable and unverifiable claims is reliable for the verifiable class (.82 \F). Detecting unverifiable claims is more challenging (.27~\F).

\section{Annotation study}
We design an annotation study with two goals: (1) To construct a resource that enables us to explore claim properties, i.e., the role of entities and relations in the fact verification process. (2) To observe how fact-checkers modify entity-based search queries during the evidence retrieval process.

We construct a dataset with two annotation layers: (a) fact
verification annotation, i.e., claims checked against evidence, and
(b) structured knowledge, i.e., entity and relations. We build our
dataset on top of \bear, a corpus of English tweets
annotated with \underline{b}iomedical \underline{e}ntities \underline{a}nd \underline{r}elations.

\subsection{Data}
\label{data-sampling}
We identify relevant claims from \textsc{Bear} for further annotation. The tweets have to:
\begin{compactitem}
    \item[\textbf{Contain a claim.}] To identify tweets that convey a claim we employ a pretrained claim detection model \citep{wuhrl-klinger-2021-claim} and only keep instances with claims. 
    \item[\textbf{Contain at least one medical relation.}] We use the documents for which at least one medical relation was annotated.
\end{compactitem}

Out of 2,100 documents from \textsc{Bear} this filtering leaves us with 646 claim-containing documents.
To extract claims, the tweets undergo two steps:
\begin{compactitem}
    \item[\textbf{Claim extraction.}] We identify individual claims within the tweets by extracting an entity--relation--entity triplet from the input documents based on the entity--relation annotation.
    \item[\textbf{Manual filtering.}] To ensure data quality, we remove 166 claims that are incorrectly extracted from the tweet's context, repetitions within the same tweet, contain the relation ``somehow related to", or are off-topic (e.g., discuss treatments of animals)\footnote{Table \ref{tab:filtering-criteria} shows a description of each category.}. 
\end{compactitem}
 
We correct grammatical errors in 346 of the automatically extracted claims to increase their readability. Table~\ref{tab:data-filtering-example} shows an end-to-end example of the filtering process.
These preprocessing steps lead to 1,532 claims to be fact-checked.

\subsection{Annotation}

\subsubsection{Annotation Task}
\label{annot-task}
In the annotation study, annotators are tasked to verify claims against scientific evidence. For every claim, their task is to find an article which contains supporting or refuting evidence for the claim. They search for evidence using PubMed\footnote{\url{https://pubmed.ncbi.nlm.nih.gov/}}, a database for biomedical articles. Based on the evidence they find, the claims are assigned a fact-checking verdict.
Claims can thus be labeled as follows:

\begin{compactitem}
    \item[\textsc{Supported}] Evidence supports the claim.
    \item [\textsc{Partially Supported}] Evidence partially support
      the claim, e.g., if evidence is more specific than the claim.
    \item[\textsc{Partially Refuted}] Evidence refutes the claim but is more specific than the claim.
    \item[\textsc{Refuted}] Evidence refutes the claim.
\end{compactitem}

We provide the annotators with a starting query for the evidence search. The query is made up of the medical entities mentioned in the claim, connected by an \textsc{and} operator. For example, a claim stating that ``H2 blocker treats SpO2'' has the starting query \href{https://pubmed.ncbi.nlm.nih.gov/?term=(H2%20blocker)+AND+(SpO2)}{``(H2 blocker) AND (SpO2)''}.

Annotators provide the PubMed Identifier (PMID) of the respective article that they use to verify the claim along with the sentences that support or refute the claim. 
For a given claim, we instruct the annotators to go over the titles and abstracts of the first five search results for the pre-built query. After that, annotators refine the search query to fine-tune their search. If the refinement leads to evidence being discovered, we record their updated search query. 
If no evidence is discovered after three minutes, the claim is labeled as \unverif.\footnote{Note that this time limit only affects the refinement process, not the overall annotation process for an instance.} To understand why a particular claim appears to be unverifiable, annotators rate how confident they are that a continued search could uncover evidence. We refer to this as the `evidence exists confidence'\footnote{Annotators rate this on a 5-point scale ranging from \textit{I'm very confident relevant evidence exists} to \textit{Very sure that there is no evidence out there which I could use to check the claim}.}.

\subsubsection{Evaluation}
\label{annotation-eval-metrics}
We evaluate the agreement for the two subtasks as follows: For the verdict assignment task we report the Cohen's $\kappa$ score between annotators.
For evidence retrieval, we gauge how often two annotators retrieve the same evidence document to verify a claim. We report the Jaccard similarity between the set of evidence documents which, given two sets, is calculated by dividing the size of the intersection of the two sets by the size of their union. Scores $>$ 0 indicate that there is at least one shared member in the sets. We calculate the Jaccard similarity for each pair of evidence documents where both annotators assigned the same verdict. Since annotators may use the same evidence document to substantiate conflicting verdicts (e.g., A1 uses document 123 and assigns \supp, while A2 uses document 123 to \refu a claim), we also report the Jaccard score for conflicting verdicts.

\subsubsection{Annotation Procedure}
\label{annot-procedure}
We set up the study using the online platform SoSci~Survey\footnote{\url{https://www.soscisurvey.de/}}. We provide screenshots of the environment in the supplementary material\footnote{The data and supplementary material is available at \url{http://www.ims.uni-stuttgart.de/data/bioclaim}.}.
We work with two in-house annotators (A1, A2) to label the claims. The annotators are male and female, aged 25 to 30, with backgrounds in computational linguistics. While they have no formal medical training, they are experienced annotators for biomedical social media data.

Both annotators label a test batch of 10 claims to evaluate our annotation guidelines.
For assigning the FC verdict, Cohen's $\kappa$ is $1.0$, indicating perfect agreement. For the evidence retrieval task, we calculate the Jaccard similarity between the evidence documents. The average similarity in all pairs where annotators assigned \supp is 0.29. Note that annotators do not assign \refu to any claim in the test batch, hence why we can not report a Jaccard score. For the  SUP--SUP instances, only in 29 \% of cases, annotators used the same evidence document to reach their verdict. This is noteworthy, as they are in perfect agreement about all verdicts. Our observation points to a key property of fact-checking evaluation: annotators may use different evidence documents to verify a claim. Low Jaccard scores therefore do not necessarily indicate low annotation quality.

In the main annotation phase, A1 and A2 research a total of 1,448 claims.\footnote{Due to limited resources,  annotators could not complete the annotation for all 1,532 claims available in the dataset.} We refer to the dataset as \bearfact.

\subsection{Corpus statistics}
For a better understanding of the resulting dataset, the following section provides corpus statistics.
\begin{table}
    \centering \small
    \begin{tabular}{lrrrrrr}
        \toprule
        claims & \textsc{Un} & \textsc{REF} & p\textsc{REF}  & p\textsc{SUP} & \textsc{SUP}& total \\
        \cmidrule(lr){2-2}\cmidrule(lr){3-3} \cmidrule(lr){4-4}\cmidrule(lr){5-5} \cmidrule(l){6-6} \cmidrule(lr){7-7}
        \# & 447 & 60 & 38 & 224 & 679 & 1448\\
        \% & 30.9 & 4.1 & 2.6 & 15.5 & 46.9 &100\\

        \bottomrule
    \end{tabular}
    \caption{Distribution of fact-checking verdicts.}
    \label{tab:bear-fact-verdict-distribution}
\end{table}
\paragraph{Number of medical claims per tweets.}
\bearfact consists of 1,448 claims in 572 tweets. We provide a a histogram of the number of claims per tweet in the Appendix (Fig. \ref{fig:number-of-claims-per-tweet}). We find that 201 out of the 572 tweets convey exactly one claim each. The tweet with the highest number of claims conveys 14 claims.
 Notably, the majority of tweets in \bearfact expresses more than one claim. 

\paragraph{Distribution of fact-checking verdicts.}
Table \ref{tab:bear-fact-verdict-distribution} provides an overview of how fact-checking verdicts are distributed in \bearfact. The majority of claims are (\textsc{partially}) \supp (62.4 \%). (\textsc{partially}) \refu has the smallest number of instances (6.7 \%). Notably, 30.9 \% of claims are unverifiable, meaning that there was no evidence to substantiate a verdict.\footnote{From the annotators' comments we learn that 17 claims have been extracted incorrectly. This means that the claim we obtain from the entity-based claim extraction step (ref. to Sec.~\ref{data-sampling}) does not accurately represent the claim expressed in the tweet, for example because the extracted claims omits relevant contextual information. As they are not checkable, we assign those instances to the \unverif class.}

For the claims that are (partially) \supp by the literature we can infer that the information conveyed in them are most likely to be true. However, 37.6 \% of instances are either \unverif or (partially) \refu, meaning there is no point of reference about that statement in the literature or they convey false information.
This finding emphasizes the importance of fact-checking for biomedical claims for social media.

\begin{figure}
    \centering
    \includegraphics[scale=.6]{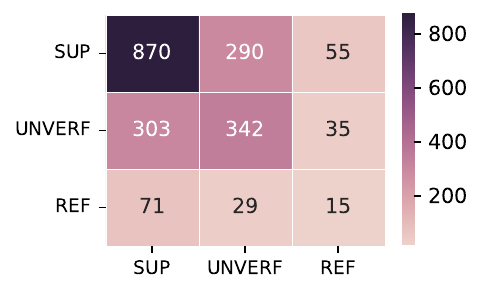}
    \caption{Pairwise co-occurrence of verdicts in \bearfact tweets
      with more than one claim. (partially) \supp and
      (partially) \refu are collapsed, resp.}
    \label{fig:verdict-cooccurrence}
\end{figure}

\paragraph{Verdict co-occurrence.}
For tweets with more than one claim, we analyze if medical claims co-occur with specific other types of claims with respect to their fact-checking verdict. In other words, if a claim in a tweet is \supp, \refu or \unverif, how common is it for the other claims in that tweet to have the same verdict?
For all tweets with more than one claim, we form all possible pairs of claims in one tweet to obtain the verdict co-occurrences. We choose pairs to be able to handle the varying amount of claims per tweet. Subsequently, we visualize the pairwise co-occurrence of their fact-checking verdicts in Figure~\ref{fig:verdict-cooccurrence}.
We observe two major patterns: In the diagonal, we see that (\textsc{partially}) \supp claims most frequently co-occur with claims of the same verdict, followed by pairings with \unverif claims.
Further, (\textsc{partially}) \refu claims do not show this pattern. Pairs of (\textsc{partially}) \refu claims are very infrequent; it is more common for such claims to occur with (\textsc{partially}) \supp or \unverif claims. 

Our analysis indicates that medical tweets have a tendency to convey claims with mixed veracity levels emphasizing the importance of a fine-grained approach to fact verification.

\section{Claim characteristics \& evidence discovery}
\subsection{Which properties make claims (un)verifiable? (RQ1a)}
\label{claim-characteristics-analysis}

\begin{figure*}
     \centering
     \begin{subfigure}[b]{0.3\textwidth}
         \centering
         \includegraphics[scale=0.47]{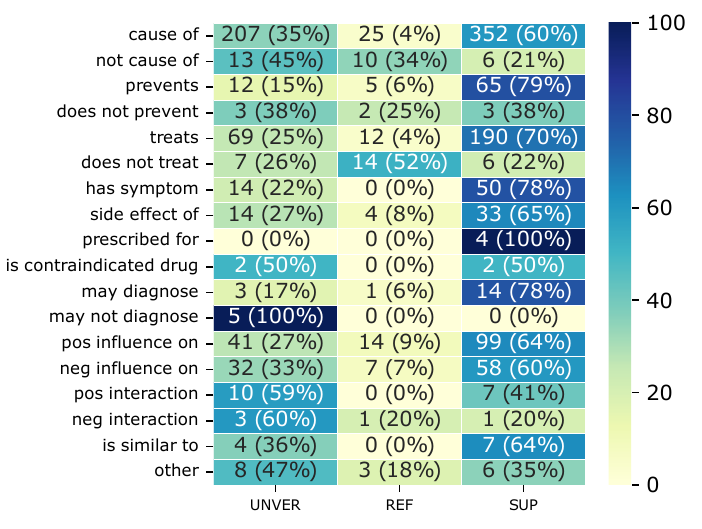}
         \caption{Relation types.}
         \label{fig:verdict-distribution-across-claim-relation}   
     \end{subfigure}
     \hfill
     \begin{subfigure}[b]{0.3\textwidth}
         \centering
         \includegraphics[scale=0.5]{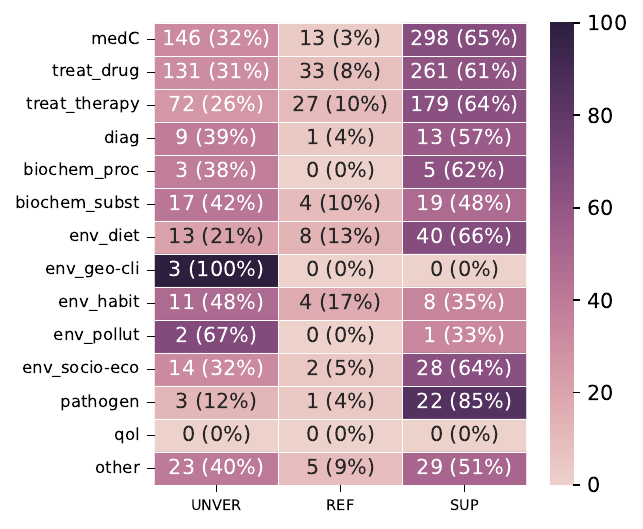}
         \caption{Subject entities.}
         \label{fig:head-entities}
     \end{subfigure}
     \hfill
     \begin{subfigure}[b]{0.3\textwidth}
         \centering
         \includegraphics[scale=0.5]{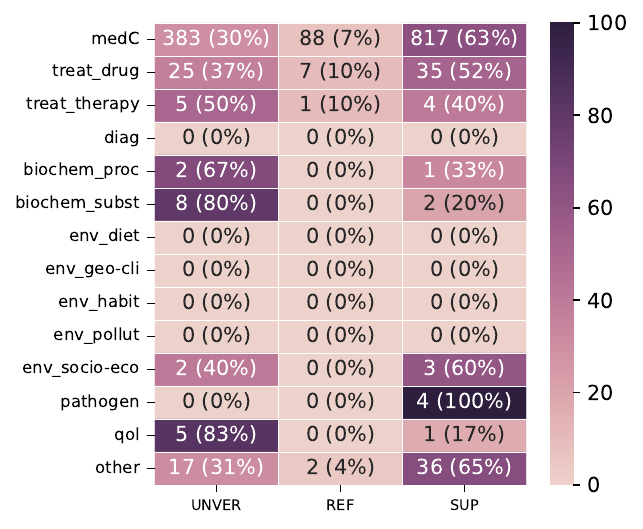}
         \caption{Object entities.}
         \label{fig:object-entities}
     \end{subfigure}
     
        \caption{Verdict distribution across claim relation and entity types. Color coding is based on the percentage of verdicts per relation/entity class. We collapse (partially) \supp and (partially) \refu instances into one group, respectively.}
        \label{fig:verdict-distr-relations-entities}
\end{figure*}

We hypothesize that entity--relation properties in a claim are connected to its verifiability. To investigate this, we explore the following questions:

\paragraph{Which claim relations are (un)verifiable?} To investigate if there are specific relation types in claims that tend
to be \supp, \refu or \unverif, we analyze the distribution of
verdicts across each relation type. Figure
\ref{fig:verdict-distribution-across-claim-relation} shows the
results. Each row in the heatmap represents a relation class, each
column represents a fact-checking verdict. Each cell depicts the
percentage of claims that express the relation and verdict.

We observe that for the positive relations, e.g., \textit{cause--of}, \textit{prevents} or \textit{positive--influence--on} the distribution is dominated by the overall verdict distribution within \bearfact, meaning the majority of claims fall into the \supp category, followed by \unverif and \refu. However, for their negative counterparts, e.g., \textit{not--cause--of}, \textit{does--not--prevent} etc., we observe the opposite. The dominant verdict class is \refu or \unverif. The exception is the pair \textit{positive/negative influence on} where the distribution of the negative relation is very similar to its positive counterpart.

To test if the distribution of positive and negative variants of a relation are in fact different, i.e., there is no relation between the two distributions, we compute the chi-square statistic\footnote{We use the Scipy implementation of chi-squared: \url{https://docs.scipy.org/doc/scipy/reference/generated/scipy.stats.chi2_contingency.html}} for relation classes with $>$ 5 instances (\textit{(not--)cause--of}, \textit{(does--not--)treat}, \textit{pos./neg.--influence--on}). The results show that for two out of three relation pairs, i.e., \textit{(not--)cause--of} and \textit{(does--not--)treat}) the distributions are in fact unrelated (p$<$0.05).

\paragraph{Which claim entities are (un)verifiable?} Figures \ref{fig:head-entities} and \ref{fig:object-entities} illustrate the distribution of entity types across each fact-checking verdict. From Fig. \ref{fig:head-entities}, we observe that the vast majority of claims use a medical condition (\textit{medC}) or treatment mention (\textit{treat\_drug}, \textit{treat\_therapy}) as their subject. The distribution across the verdicts for the majority of entity classes mirrors the distributing of verdicts in the overall dataset.
In Fig. \ref{fig:object-entities}, we observe that claim objects are almost exclusively mentions of medical conditions.
This is most pronounced in (\textsc{partially})\refu claims where 88 out of 98 (\textsc{partially})\refu claims make a claim. Across all verdicts, environmental factors are almost exclusively claim objects.
When considering the entity distributions, note that to a certain degree the relation type dictates the subject and object entity type in a triplet.

\subsection{Evidence discovery: Are medical entities meaningful search queries? (RQ1b)}
\label{evidence-discovery-analysis}

We investigate \textbf{how we can use medical entities in the
  claims as meaningful search queries for evidence discovery}. We
hypothesize that medical entities are key to connect claims to
evidence. Further, we explore how search queries are refined during
the evidence retrieval process. Thus, we analyze our annotators' search strategies. 

\paragraph{Are medical entities meaningful search queries?}
Recall that annotators start their search with pre-built search queries that consist of the medical entities from the claim connected by \textsc{and} operators (see Sec.~\ref{annot-task}).
Out of the 1,001 claims in our dataset for which annotators found supporting or refuting evidence, 757 claims could be verified with the results from this original search query. In 244 cases (24.4 \%) the annotators had to refine the search query, and subsequently discovered an evidence document.
This shows that medical entities are an appropriate starting point for evidence search.\footnote{Note that PubMed's internal article ranking contributes to evidence discovery. The properties of this ranking need to be taken into account when designing systems that do not rely on PubMed.}

\paragraph{How are queries refined to discover suitable evidence?}
We aim to understand how search queries are refined and analyze the query refinement strategies that lead to evidence being discovered.
To that extend, we define seven types of refinements:
\begin{compactenum}
    \item Generalizing search terms
    \item Specifying search terms
    \item Normalizing brand names by replacing brand names with the respective active ingredient
    \item Normalizing informal language
    \item Resolving abbreviations 
    \item Adding relation between search terms
    \item Other
\end{compactenum}

Table \ref{table:refinement} shows examples and the number of search terms per refinement type class in a subsample of 50 claims. In those claims, annotators refined a query and subsequently discovered an evidence document. We count a total of 56 refinement operations in our sample. Normalizing informal and colloquial terms is the most frequent operation (in 18 out of 50 instances), followed by using a more general search term (11) and resolving abbreviations (10). Adding a relational term to the query is used the least (2).

With the exception of Strategy 6 (Adding relation), all other strategies we observe are operations to normalize the query terms. Considering the style gap between the social media claims and the scientific evidence documents, this finding intuitively makes sense. It also indicates that entity normalization or linking has the potential to improve automatic evidence retrieval methods.

\begin{table*}[t]
  \centering \small
  
  \begin{tabularx}{\linewidth}{rXrXX}
    \toprule
    Id & Refinement type &\#refined& Original term/query & Refined term/query \\
    \cmidrule(lr){1-1}\cmidrule(lr){2-2} \cmidrule(lr){3-3} \cmidrule(lr){4-4} \cmidrule(lr){5-5}
    1 & Generalizing search term &11 &(Vit C 500mg) &(Vitamin C) \\
    2 & Specifying search term & 8&(delta) &(delta) AND (Covid)\\
    3 & Normalizing brand name&4 &(Tecentriq)& (Atezolizumab)\\
   4 & Normalizing informal terms& 18 &(Rona)& (corona) \\
   5 & Resolving abbreviations& 10& (IBS) &(irritable bowel syndrome) \\
   6 &Adding relation &2&(fever) AND (Ivermectin)&(fever) side effect of (Ivermectin)\\
   7& other &3&(elevated Uric acid levels)& (Uric acid) AND (lower)\\
   
    \bottomrule
  \end{tabularx}
  \caption{Number of refined search terms per refinement type across a sample of 50 successfully refined search queries along with examples.}
  \label{table:refinement}
\end{table*}

\paragraph{Why do claims remain unverifiable?}
447 claims are labeled as \unverif, because the query refinement was
not successful and no relevant evidence could be discovered. To
understand why a particular claim could not be checked, we analyze the
annotators' estimate that evidence exists.
For the majority of cases (54.1 \%), annotators state that they
cannot judge if evidence could exist. However, approx.\ 20 \% of the
unverifiable claims, the annotators are confident that evidence exists
(\textit{pretty confident}: 15.9\%, \textit{very confident:
  4.7\%}). For the remaining 25\% of \unverif claims, annotators are
either \textit{pretty} (15.2\%) or \textit{very} confident (8.9\%)
that no evidence exists (Ex.\ 1/2 in Table~\ref{tab:no-evidence-examples} in Appendix). Those claims are ambiguous or general in which case it makes sense that discovering evidence is unlikely. 
For the claims with high confidence about the existence of evidence, consider Examples 3 and 4 in Table \ref{tab:no-evidence-examples}.

\section{What is the impact of the annotation setup?}
\label{crowd-settings}
As described in the previous section, we employ in-house annotators to
create \bearfact. Crowdsourcing is, however, a viable alternative to
collect fact-checking annotations \citep{martel2023,
  mohr-etal-2022-covert}. Therefore, to understand how the annotation
setting, i.e., in-house annotation vs.\ crowdsourcing, impacts our
task, we investigate \textbf{RQ2a: }\textbf{How reliably do untrained
  crowdworkers verify biomedical claims?} and \textbf{RQ2b What is the
  impact of domain knowledge} (i.e., biomedical expertise) \textbf{in
  the crowdsourcing setting?}

\subsection{Experimental setting}
To study this we require crowd annotations from people with and
without biomedical expertise.  For the general crowd (to which we
refer as layCrowd), we recruit university students of a Master's
Program in Computational Linguistics for a voluntary on-site
study. For the crowdworkers with domain expertise (expCrowd), we
recruit participants with a background in (bio)medicine on the online
crowdsourcing platform Prolific\footnote{\url{https://prolific.com}}.
Each participant verifies ten claims. The study is hosted using Google
Forms\footnote{\url{https://docs.google.com/forms/}}. Crowdworkers on
Prolific are reimbursed with \textsterling9 per hour. All partipants
are instructed with the same guidelines as in-house annotators.

\subsection{Results}
\subsubsection{Fact-checking verdicts}
We obtain ten sets of annotations in the expCrowd setting and nine
sets in the layCrowd setting\footnote{One annotator in the layCrowd
  setting dropped out.}.  Table \ref{table:iaa} presents the agreement
scores for the verdict assignment task.\footnote{One participant in
  layCrowd only managed to work on 8 out of 10 claims in the scheduled
  time. We include their annotations for the completed claims. For the
  agreement, we calculate the $\kappa$ metric for all pairs involving
  this annotator only with the completed claims.} We report the
agreement among each group (layCrowd, expCrowd) as well as the
agreement between our in-house annotations and the aggregated label
from each group. Assuming that for modeling purposes we would use an aggregation of the individual crowdworkers' labels, we first aggregate via majority voting
before calculating the agreement between groups. Appendix
\ref{appendix:aggregation} shows details on the aggregation strategy.

\begin{table}
  \centering\small
  \begin{tabular}{llr}
    \toprule
    Annotator 1& Annotator 2 &$\kappa$  \\
    \cmidrule(r){1-1} \cmidrule(lr){2-2}\cmidrule(l){3-3} 

    \multicolumn{2}{c}{layCrowd} & 0.24 \\
    \multicolumn{2}{c}{expCrowd} & 0.22 \\
    \cmidrule(r){1-1} \cmidrule(lr){2-2}\cmidrule(l){3-3} 

    in-house& agg. layCrowd & 0.30\\
    in-house& agg. expCrowd & 0.40 \\ 
    agg. expCrowd & agg. layCrowd &0.65\\
    \bottomrule
  \end{tabular}
  \caption{Inter-annotator agreement for the verdict assignment task. We report (av.) pairwise Cohen's $\kappa$.}
  \label{table:iaa}
\end{table}

\paragraph{Trained annotators compared to crowdworkers.} 
The agreement among the individual annotators in layCrowd is
$\kappa=0.24$, indicating fair agreement. The result is similar for
expCrowd ($\kappa=0.22$). The in-house annotators of the full corpus
showed perfect agreement in their training phase. These scores are not
directly comparable, but this indicates that the task is more
challenging in a crowd setting.  The agreement between the two crowd
settings (agg.\ expCrowd, agg.\ layCrowd) is moderate ($\kappa=0.65$).

\paragraph{Impact of biomedical expertise.}
To understand if domain expertise has an impact on the annotation, we
compare the agreement scores for layCrowd and expCrowd. The agreement
among the domain experts is slightly lower (a $\Delta$ of
$0.02\kappa$). This indicates that biomedical expertise does not have
a substantial impact on the reliability of the results, in fact, their
judgments are more varied compared to the general crowd. Note,
however, that annotation quality in an anonymous online setting may
also vary more compared to an on-site setting.

The agreement between the in-house annotators and the agg.\ expCrowd
verdicts is higher than their agreement with the agg.\ layCrowd
verdicts. We hypothesize that this may be an effect of their prior
experience in annotating biomedical data.

Finally, we visualize the verdict assignments in a confusion matrix in Table~\ref{tab:conf-matrix} which shows the verdicts assigned by agg.\ layCrowd on the vertical and the agg.\ expCrowd  verdicts on the horizontal axis. The diagonal represents the instances where both groups assigned the same verdict. We observe the strongest agreement in the \unverif instances. Notably, we observe that in two instances the agg.\ expCrowd is able to verify the claim (\supp), the lay crowd is not. We presume that this a result of their domain expertise.

\begin{table}[]
    \centering \small
    \begin{tabular}{llccc}
    \toprule
   & &\multicolumn{3}{c}{expert}\\
   \cmidrule(lr){3-5}
   &&  SUP & UNV & REF \\
       \multirow{3}{*}{\rotatebox[origin=c]{90}{lay}} & SUP &3  &0 & 0  \\
        & UNV & 2 & 5 & 0 \\
        & REF &0 & 0 &0  \\

        \bottomrule
    \end{tabular}
    \caption{Confusion matrix illustrating the verdicts assigned by agg.\ layCrowd vs. agg.\ expCrowd.}
    \label{tab:conf-matrix}
\end{table}

\subsubsection{Evidence retrieval}
We want to understand how frequently annotators use the same evidence
to \textsc{support} or \textsc{refute} a claim, while the verdict
label is or is not the same. Table
\ref{tab:evidence-retrieval-jaccard-crowd} reports the Jaccard
similarities that measure the overlap of evidence documents. For each
combination of verdicts (SUP--SUP, REF--REF, SUP--REF), we report the
average Jaccard score as well as the number and percentage of
instances within that group with a Jaccard score $>$0.  The upper half
of the table reports the results for layCrowd, the bottom half the
results for expCrowd.

Across all verdict combinations, layCrowd annotators agree more on
which PubMed document they use to substantiate a verdict compared to
the expCrowd annotators.  For the SUP--SUP instances, in 48~\% of
evidence pairs, layCrowd annotators relied on at least one common
document, while in expCrowd this is only the case for 33 \% of
evidence pairs. Apparently, experts chose the evidence more
selectively and do not accept the first document that might be a
fit. This is in line with the agreement scores for the verdict
assignment: If annotators use more diverse evidence documents, we
can also expect their verdicts to be more diverse.
Generally, the agreement on evidence documents is the highest in
REF--REF pairs compared to the other combinations of verdicts. This
may be an effect of negative results being published more seldomly.

Annotators sometimes use the same evidence documents, but reach opposing
verdicts. This happens more frequently in the layCrowd setting. For
SUP--REF pairs, we observe an average Jaccard score of $0.38$ in
layCrowd and $0.3$ in expCrowd. We hypothesize that this is an effect
of domain knowledge as it might take biomedical expertise to
interpret evidence correctly.

The relatively low Jaccard scores potentially also result from the annotators' evidence research strategy. In cases where the query returns multiple relevant evidence documents, one annotator may stop their research after discovering the first document, while the other continues, the Jaccard score is zero. In our setup, we do not explicitly instruct annotators to look at all evidence documents before assigning the verdict. Instead, we instruct them to return to the survey once they find suitable evidence. While we assume that they go over the results from top to bottom, we cannot control for that. Defining more fine-grained retrieval strategies could be an interesting task for future research.
That being said, we do not strictly want to optimize for a high Jaccard score. Two annotators could reach the same verdict using different evidence documents, so we should always take into account both the agreement for the verdict assignment as well as the Jaccard score for evidence retrieval.

\begin{table}
    \centering \small
    \begin{tabular}{rrrrr}
        \toprule
         && SUP-SUP & REF-REF & SUP-REF \\
         
        \cmidrule(r){2-2}\cmidrule(lr){3-3} \cmidrule(lr){4-4}\cmidrule(l){5-5}
        \multirow{3}{*}{\rotatebox[origin=c]{90}{lay}} &avg. J & .31 & .58& .38 \\
        &\# J$>$0 & 40&10 &27\\
        &\% J$>$0 &.48 & .77 &  .56 \\
        
        \cmidrule(r){2-2}\cmidrule(lr){3-3} \cmidrule(lr){4-4} \cmidrule(l){5-5}
        \multirow{3}{*}{\rotatebox[origin=c]{90}{expert}}&avg. J & .19& .43.&.2\\
        &\# J$>$0 &53 &4 &17\\
        &\% J$>$0 &.33 &.57 &.3\\

        \bottomrule
    \end{tabular}
    \caption{Avg. Jaccard similarity (avg. J), number and percentage of Jaccard scores $>$ 0 per verdict combination for the lay and expert crowd. We collapse the partially \supp/\refu verdicts into their respective major class.}
    \label{tab:evidence-retrieval-jaccard-crowd}
\end{table}

\section{Can we predict if claims are unverifiable?}
\label{modeling}
Moving to the modeling perspective, we investigate if we \textbf{can predict verifiability, i.e., the likelihood that evidence for a claim exists, purely from the claim}
(RQ3). A setup like that could allow us to adapt the manual evidence
search procedure by giving experts more time for such claims. To
investigate this, we train a model that differentiates \unverif from
verifiable (\supp/\refu) claims.

\paragraph{Experimental setting.} To train the models, we define two classes: \unverif and {\textsc{verifiable}\xspace} ((\textsc{Partially}) \supp, (\textsc{Partially}) \refu). We train a classifier on top of RoBERTa \citep{conneau-etal-2020-unsupervised} for 15 epochs with default parameters on an Nvidia RTX A6000, using a 80/20 train-test split of the \bearfact data. The input for the classifier is the claim phrase.

\paragraph{Results.} Table \ref{tab:classifiers_scores} shows the results. They
indicate that it is possible to reliably identify claims that
are verifiable (.82\F), whereas identifying unverifiable claims proves more challenging (.27\F). 
We further look into the connection between the model performance and annotators' `evidence exists confidence' (see Sec.~\ref{annot-task}). We hypothesize that this rating may be correlated with the predicted labels. We find, however, that the point biserial correlation\footnote{\url{https://docs.scipy.org/doc/scipy/reference/generated/scipy.stats.pointbiserialr.html}} that measures the correlation between the binary predicted labels and the continuous confidence scores is limited (0.22).

\paragraph{Qualitative Analysis.} To expand our analysis of
(un)verifiable claims to the modeling perspective, we conduct a manual
error analysis.  We find that {\textsc{verifiable}\xspace} claims frequently
include technical terms and medical terminology making the entities
and relation more specific (see Ex.\,1,
Table~\ref{table:error-analyis-examples}). \unverif claims on the other hand
include more general vocabulary, potentially making them more
difficult to verify (see Ex.~2). Based on this introspection, we
hypothesize that the model picks up on the varying specificity levels
of the medical terminology and overgeneralizes this property in the
incorrectly classified instances (Ex.~3, 4). This aligns with
Sec. \ref{evidence-discovery-analysis} where we find that entity
normalization is key to verify claims.

\begin{table}[t]
  \centering \small
  
  \begin{tabularx}{\linewidth}{rXrr}
    \toprule
    Id & claim &G& P \\
    \cmidrule(r){1-1}\cmidrule(lr){2-2} \cmidrule(lr){3-3} \cmidrule(l){4-4}
     1& dantrolene treats malignant hyperthermia & V & V\\
     2& inappropriate cleaning methods is/are the cause of outbreaks & U & U\\
     3& NAC is contraindicated drug to pregnancy & U & V\\
     4& fear causes a weakened immune system & V & U\\

    \bottomrule
  \end{tabularx}
  \caption{Claims along with gold and predicted verifiablitiy labels
    (U: \unverif, V: \textsc{verifiable}). G: gold, P: Prediction.}
  \label{table:error-analyis-examples}
\end{table}

\begin{table}
    \centering \small
    \begin{tabular}{rrrrr}
        \toprule
         Class & Recall & Precision & F1 \\
        \cmidrule(r){1-1} \cmidrule(lr){2-2}\cmidrule(lr){3-3} \cmidrule(l){4-4}
        {\textsc{verifiable}} & .93 & .74 & .82 & \\
        \unverif & .18 & .52 & .27 & \\
        \cmidrule(r){1-1} \cmidrule(lr){2-2}\cmidrule(lr){3-3}\cmidrule(l){4-4}
      Macro & .56 & .63 & .54 \\
      Weighted & .71 & .67 & .66 \\
        \bottomrule
    \end{tabular}
    \caption{Precision, Recall, \F of the verifiability class.}
    \label{tab:classifiers_scores}
\end{table}

\section{Related Work}

\textbf{Scientific biomedical fact-checking} focuses on verifying scientific claims against evidence sources. Typically, this is modeled in two steps: evidence retrieval, i.e., discovering relevant sources, and claim verification, the task of assigning a verdict to a claim based on the evidence. This can also be modeled jointly. \citet{vladika-matthes-2023-scientific} provide a comprehensive overview.

The focus in the area has been on concise, sometimes synthetic claims
\citep{wadden-etal-2022-scifact,kotonya-toni-2020-explainable-automated}. More
recently, user-generated medical content has received more attention
\citep[i.a.]{zuo-et-al_2022, vladika2023healthfc,
  saakyan-etal-2021-covid}.  This type of content is challenging
\citep{kim-etal-2021-robust} and complex
\citep{sarrouti-etal-2021-evidence-based}, posing the question of
which units of information should serve as the input to FC
systems. Nevertheless currently it is standard to either process full
sentences or even paragraphs
\citep{mohr-etal-2022-covert}. Alternatively claims are atomic by
design, e.g., claims in \textsc{SciFact}
\citep{wadden-etal-2022-scifact}.
Recent work shows that both evidence retrieval and claim verification are more robust for concisely formulated claims. Apart from that, \textbf{properties of biomedical claims} and their impact on fact-checking are poorly understood.

Outside biomedical fact-checking, i.e., in argument mining and
argument theory, there exists a more developed understanding of claim
properties. In argument mining claims have been categorized
according to their function, i.e., epistemic vs.\ practical vs.\ moral
claims \citep{Lippi_Torroni_2016}, their semantic type
\citep[i.a.]{hidey-etal-2017-analyzing,egawa-etal-2019-annotating,jo-etal-2020-machine}
or studied with respect to their conceptualization across domains
\citep{daxenberger-etal-2017-essence,boland2022}.

For fact-checking, some datasets categorize their claims into groups such as numerical claims or position statements \citep{fast-furious-FC-challenge}, but the inherent structural properties of claims are not understood. Structured knowledge has been proposed to represent claims in scientific discourse \citep{magnusson-friedman-2021-extracting}, and as a method to detect \citep{yuan-yu-2019} and extract health-related claims \citep{wuhrl-klinger-2022-entity}.

Two strands of research are related to our task of estimating a claim's (un)verifiablity: studies that explore stylistic properties of claims to detect misinformation \citep[i.a.]{rashkin-etal-2017-truth,schuster-etal-2020-limitations} and \citet{atanasova-etal-2022-fact} who detect when evidence with omitted information is (in)sufficient to reach a fact-checking verdict.

\section{Conclusion}
With this paper, we contribute a better understanding of what makes
claims (un)verifiable. To this end we design a study that tasks
annotators with varying levels of expertise to search for evidence and
label claims that have entity/relation annotations. In an in-depth analysis of the resulting resource we find that claims with particular relations are more challenging to find evidence for and successfully assign fact-checking verdicts. \emph{This leads to
  important future work, namely to focus on methods that are able to
  find evidence also for negated claims (X--does-not-cause--Y).}
Through the study we also observe that some specific topics appear
to be more challenging, including environmental factors. \textit{We
  suggest that future work studies which data bases are promising
  sources to provide evidence for specific biomedical
  topics. Presumably, PubMed is not equally well suited across
  topics.}

We further analyze if the expertise level of annotators has an impact
on the annotation quality and how far the annotations by various
groups overlap. Evidently, domain expertise leads to more
carefully selected evidence (which might also be more
accurately selected), but not to a too large difference in verdicts. \textit{This aspect requires further studies -- how exactly does an evidence document need to relate to a claim to allow for
a correct verdict? Does, perhaps, the inference procedure vary
between annotators for a good reason? Following a perspectivist approach, the diversity in verdicts should be carefully investigated in future studies.}

Finally, we perform a modeling study to understand if we can develop
a system supporting the annotation process. We find that a text
classifier is surprisingly successful to predict if a claim is likely
to have evidence -- with a nearly perfect recall and a high
precision. \textit{Future work is required to study how such
  classifier can be used in an annotation setup. We assume that not
  restricting the time of annotators to find evidence for such claims
  would be an appropriate design decision.} Apart from that, future work should explore other modeling approaches such as few-shot prompting to explore the capabilities of LLMs for our task. With respect to both understanding claim and evidence properties as well as modeling, we further need to explore how to handle complex claims that may not follow the subject-relation-object structure we are investigating in this work.

\section{Ethical Considerations}
It lies in the nature of fact verification that annotators may be exposed to false medical information. We educate annotators about this possibility before they start the task. They can stop working on the task at any time.
The resulting resource is first and foremost a dataset intended for analyses, designed to enable further research in modeling biomedical claim verification. It should not be the basis of in-production, fully automatic fact-checking systems. Further, biomedical research itself constantly evolves which means the evidence and verdicts in \bearfact may be outdated.

It is important to point out that the fact-checking verdicts in the dataset are not to be taken out of context from the evidence document. While we can cautiously infer a veracity label for claims that are \supp or \refu, because we trust annotators to base their verdict on reliable evidence, there is no objective measure of truth in our task. In theory, there could be a number of reasonable evidence documents for a single claim, as medical research might produce multiple studies on the same topic.

\section{Limitations}
Our work studies claim properties based on real-world data. Therefore,
the conclusions we draw from the analyses represent the underlying
sample. The extend to which our findings generalize to other datasets
should be the focus of future research. Such a manually annotated
corpus can never represent the entirety of data in the real-world
appropriately. While we believe that the corpus we created can be used
to induce machine learning models, it might lead to biases and other
unwanted confounding variables, given such limitations.

While we recruit a substantial number of annotators in the crowd
experiments (§\ref{crowd-settings}) (9 and 10 crowd workers,
respectively, who work on the same claims), the number of instances we
study is comparably small. Particularly difficult or straight-forward
claims could have a stronger impact on annotation performance in the
small study setting. It is important to contextualize our findings
further. We see this as an opportunity for future work.

With respect to the evidence retrieval process, and the conclusions we
draw from annotators not being able to discover evidence for a given
claim, we have to consider that the PubMed knowledge base could be
incomplete. Evidence could indeed exist, but not being indexed by
PubMed. Further, the database does not guarantee that the documents
within it are accurate or of reliable quality.

Even with thorough document filtering (§\ref{data-sampling}), we
cannot guarantee that tweets' authors always intent to make a claim as
opposed to opinion statement. The latter usually are not considered to
be verifiable \cite{merpert-et-al_2018}. However, following
\citet{toulmin_2003} who defines a claim as an ``assertion that
deserves our attention'' \citep{toulmin_2003} we argue that in the
medical context, we need a to adopt a wide definition of what
constitutes a claim. Any statement that conveys false medical
information poses immediate harm and therefore deserves fact-checkers'
attention.

\section*{Acknowledgments}
This research has been conducted as part of the FIBISS project which is funded by the German Research Foundation (DFG, project number: KL~2869/5-1). We thank the reviewers for their valuable feedback and our annotators for their hard work and attention to detail.

\bibliography{anthology,custom}
\bibliographystyle{acl_natbib}


\appendix

\section{Appendix}
\label{sec:appendix}

\subsection{Data Processing}
Table \ref{tab:filtering-criteria} shows the filtering criteria for removing irrelevant claims before the fact-checking annotation.
\begin{table*}[h]
    \centering\small
    \begin{tabularx}{\linewidth}{XX}
        \hline
        Criterion & Description \\
        \cmidrule(lr){1-1} \cmidrule(lr){2-2}
        Incorrectly extracted from the tweet's context & Instances for which the entity-based claim extraction lead to the original statement being mis-represented, e.g., if relevant context is omitted when extracting the claim triplet. \\
        \cmidrule(lr){1-1} \cmidrule(lr){2-2}
        Repetitions & Removing claim duplicates in the same tweet. The repetions are artifacts of the annotation aggregation strategy employed in \bear. \\
        \cmidrule(lr){1-1} \cmidrule(lr){2-2}
        Contain relation ``somehow related to"& Removing claims with this relation, because they are highly unspecific and therefore not check-able. \\
        \cmidrule(lr){1-1} \cmidrule(lr){2-2}
        Off-topic claims & Claims that discuss medical conditions in animals. \\
        \hline
    \end{tabularx}
    \caption{Filtering criteria for removing irrelevant claims before the fact-checking annotation.}
    \label{tab:filtering-criteria}
\end{table*}

Table~\ref{tab:data-filtering-example} shows the filtering process described in Sec.~\ref{data-sampling} using an example.

\begin{table*}
\small
    \begin{tabularx}{\linewidth}{lXX}
    \toprule
    
        Filtering step & Method & Result for example instance\\
        \cmidrule(r){1-1}\cmidrule(lr){2-2} \cmidrule(l){3-3}
        Contains a claim & Claim classifier \citep{wuhrl-klinger-2021-claim} & True\\
        \cmidrule(r){1-1}\cmidrule(lr){2-2} \cmidrule(l){3-3}
        Contains <1 med. relation & Annotations in \bear & True \\
        \cmidrule(r){1-1}\cmidrule(lr){2-2} \cmidrule(l){3-3}
        Claim extraction & Extracting entity-based claims \citep{wuhrl-klinger-2022-entity} & `Females' negative influence on `leukopenia'\\
        \cmidrule(r){1-1}\cmidrule(lr){2-2} \cmidrule(l){3-3}
        Manual filtering & Manual inspection & passed\\
        \cmidrule(r){1-1}\cmidrule(lr){2-2} \cmidrule(l){3-3}
        Correcting grammar & Manual & Being female has a negative influence on leukopenia.\\

        \bottomrule
    \end{tabularx}
    \caption{Data filtering process exemplified with an instances from the dataset. The input tweet reads: \textit{Females tend to have greater relapses, leukopenia, more arthritis, and Raynaud phenomenon.} The claim we obtain as a result of the final filtering step is what the annotators verify during the a study. Note that we provide them with the claim as well as with the full tweet for context.}
    \label{tab:data-filtering-example}
\end{table*}

\subsection{Annotation study}

\subsubsection{Study disclaimer} 
Crowd annotators obtain the following description and task disclaimers when starting the study:

``\textbf{Purpose of this Study} We want to understand how people check if a biomedical statement is true or not and how they find evidence to judge those statements.
\textbf{Your Task} You will be presented with a Twitter post which contains a biomedical claim. You will fact-check this claim by searching for relevant evidence on PubMed, a database of biomedical articles. The study should take you about 1 hour to complete.
Please be aware that you might be looking at claims which convey false biomedical information.
You can stop the study at any time. Note that you won't be paid in this case.
\textbf{Data Collection} The data we collect will not contain any personal information. It will be used for researching automatic fact-checking and made publicly available in an anonymized form. We will write a scientific paper about this study which can include anonymized examples from the collected data."

\subsubsection{Annotator compensation}
In-house annotators are compensated with 12,52 € per hour. The crowdworkers on Prolific are compensated with \textsterling 9 per hour, which corresponds to the recommended amount on the platform.
\subsubsection{Annotator screening on Prolific}
We add the following screeners to recruit participants for the Prolific study:
\begin{itemize}
    \item Fluent languages: English
    \item Highest education level completed: Technical/community college, Undergraduate degree (BA/BSc/other), Graduate degree (MA/MSc/MPhil/other), Doctorate degree (PhD/other)
    \item Subject: Biochemistry (Molecular and Cellular), Biological Sciences, Biology, Biomedical Sciences, Chemistry, Dentistry, Health and Medicine, Medicine, Nursing, Pharmacology, Science
\end{itemize}

\subsection{Corpus statistics}
\subsubsection{Number of claims per tweet}
\begin{figure}
    \centering
    \includegraphics[scale=.45]{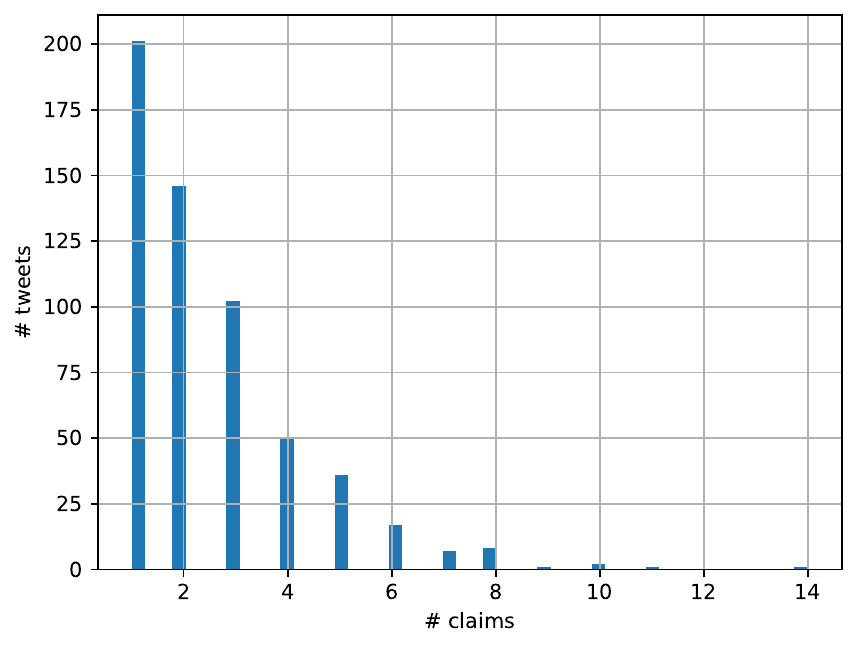}
    \caption{Number of claims per tweet in \bearfact.}
    \label{fig:number-of-claims-per-tweet}
\end{figure}

Figure \ref{fig:number-of-claims-per-tweet} visualizes the number of claims per tweet.

\subsubsection{Verdict co-occurrence}
\begin{figure}
    \centering
    \includegraphics[scale=.5]{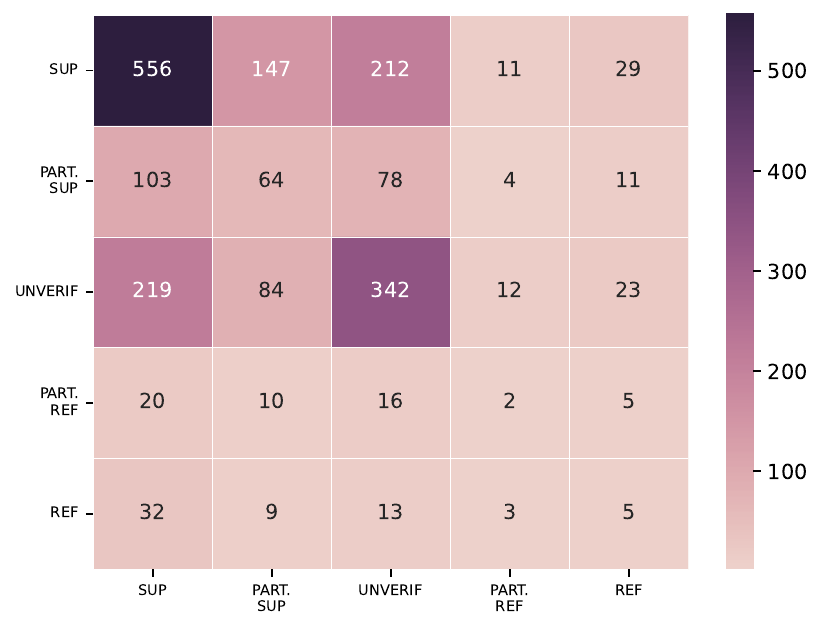}
    \caption{Pairwise co-occurrence of fact-checking verdicts in \bearfact tweets with more than one claim.}
    \label{fig:verdict-cooccurrence_fine-grained}
\end{figure}

Figure \ref{fig:verdict-cooccurrence_fine-grained} shows the pairwise co-occurrence of fact-checking verdicts in \bearfact tweets with more than one claim.

\subsection{Aggregation}
\label{appendix:aggregation}
When computing the majority vote for the fact-checking verdicts, we collapse the \textsc{partially} \supp and \textsc{partially} \refu verdicts into one group, respectively. 

\subsection{Evidence refinement}
Table \ref{tab:no-evidence-examples} shows example claims along with their confidence ratings w.r.t. if they think evidence exists and could be discovered given unlimited time and resources for research.
\begin{table}[h]
    \centering \small
    \begin{tabularx}{\columnwidth}{cXX}
        \hline
        id & Claim & Ev. Exists Confidence \\
        \cmidrule(lr){1-1}\cmidrule(lr){2-2}\cmidrule(lr){3-3}
        1 & anti-everythings are the cause of being moodier & very sure \textit{no} ev. exists \\
        \cmidrule(lr){1-1}\cmidrule(lr){2-2}\cmidrule(lr){3-3}
        2 & Strontium in chem trails is/are the cause of Cancer& pretty sure \textit{no} ev. exists \\
        \cmidrule(lr){1-1}\cmidrule(lr){2-2}\cmidrule(lr){3-3}
        3& trapped in fires is/are the cause of ptsd &pretty confident ev. \textit{ exists} \\
        \cmidrule(lr){1-1}\cmidrule(lr){2-2}\cmidrule(lr){3-3}
        4 & Magnesium glycinate 100-200mg has a pos. influence on immune system & very confident ev. \textit{exists} \\
        \hline
    \end{tabularx}
    \caption{Example claims and annotators' confidence ratings w.r.t. if evidence exists.}
    \label{tab:no-evidence-examples}
\end{table}

\end{document}